\documentclass{article}

 \usepackage{enumitem}

\PassOptionsToPackage{numbers, compress}{natbib}

\usepackage[preprint]{neurips_2024}

\usepackage[utf8]{inputenc} 
\usepackage[T1]{fontenc}    
\usepackage{hyperref}       
\usepackage{url}            
\usepackage{booktabs}       
\usepackage{amsfonts}       
\usepackage{nicefrac}       
\usepackage{microtype}      
\usepackage{xcolor}         

\usepackage{amsmath}        %

\usepackage{todonotes}
\makeatletter
\define@key{todonotes}{Konrad}[]{%
    \setkeys{todonotes}{author=@Konrad,color=green}}%
\makeatother
\makeatletter
\define@key{todonotes}{Lyle}[]{%
    \setkeys{todonotes}{author=@Lyle,color=cyan}}%
\makeatother

\title{Empirical influence functions to understand the logic of fine-tuning}

\author{%
  Jordan K.~Matelsky \\
  Department of Bioengineering \\
  University of Pennsylvania \\
  Philadelphia, PA; \\
  Research \& Exploratory Development Department \\
  Johns Hopkins University Applied Physics Laboratory \\
  Laurel, MD \\
  \texttt{matelsky@upenn.edu} \\
  \AND
  Lyle~Ungar \\
  Department of Computer and \\ Information Science \\
  University of Pennsylvania \\
  Philadelphia, PA \\
  \texttt{ungar@cis.upenn.edu} \\
  \And
  Konrad P.~Kording \\
  Department of Bioengineering, \\
  Department of Neuroscience \\
  University of Pennsylvania \\
  Philadelphia, PA; \\
  CIFAR LMB Program \\
  \texttt{kording@upenn.edu}
}

\begin{document}

\maketitle

\begin{abstract}
Understanding the process of learning in neural networks is crucial for improving their performance and interpreting their behavior.
This can be approximately understood by asking how a model's output is influenced when we fine-tune on a new training sample.
There are desiderata for such influences, such as decreasing influence with semantic distance, sparseness, noise invariance, transitive causality, and logical consistency. Here we use the empirical influence measured using fine-tuning to demonstrate how individual training samples affect outputs. We show that these desiderata are violated for both for simple convolutional networks and for a modern LLM. We also illustrate how prompting can partially rescue this failure. Our paper presents an efficient and practical way of quantifying how well neural networks learn from fine-tuning stimuli.
Our results suggest that popular models cannot generalize or perform logic in the way they appear to.
\end{abstract}

\section{Introduction}

There is great interest in understanding how the components of training sets affect neural network outputs, both to improve their performance, and to better understand their behavior in real-world scenarios.
This output is a nonlinear function of all the items in the training set, frustrating all simple interpretations.
An emerging way of thinking about the output is in terms of influence functions: every item in the training set somewhat changes the model's outputs to other inputs.
Recent work has used influence functions calculated by approximating the inverse Hessian in order to evaluate generalization, internal model state, and alignment \cite{grosse_studying_2023}.
However, the framework of empirical influence functions can be used more easily in the context of fine-tuning, where we can simply experimentally fine-tune and then demonstrate how a single training sample in isolation affects various outputs.
Importantly, due to the speed of fine-tuning on small datasets, empirical influence functions can be efficiently calculated without exceeding the compute power available to most researchers.

In this view, we can think of model training as a two step process: first, influence functions are learned, and second, these existing influence functions are used to solve a new problem. The question then is how good these systems are at applying existing data to new domains, and how much they reflect the properties that we would like a learning system to have. What would we like our influence functions to express?
More generally, what generalization properties would we want a good learner in the real world to have? The influence function of such a learner should obey the following \textbf{\textit{desiderata}} by being able to learn:

\textbf{Semantics.} Influence should generally be a decreasing function with distance in some semantically meaningful embedding space.\\
\textbf{Sparsity \& Selectivity.} The influence of low-noise data on high-noise data should be large, but not vice versa. Most samples should be non-informative about most other samples.\\
\textbf{Transitivity \& Causality.} If A causes B and B causes C then A is more likely to cause C.\\
\textbf{Ontology.} A statement like all mammals have hearts implies that all elephants have hearts but not all heart-havers are elephants.\\
\textbf{Compositionality.} If A is a part of B and B is a part of C then A is a part of C.\\
\textbf{Logical implication.} Influences should respect obvious logical statements like uncertainty quantification and negation.

In this paper, we measure empirical influence functions (EIFs) in the context of fine-tuning, while testing for these properties. We analyze both simple toy problems for which we have strong intuitions, and modern large language models, for which we are concerned with alignment and interpretability. Above all, we ask if the learning during fine-tuning, as characterized by the influence functions, obeys the desiderata of learning.
We find that fine-tuning convolutional neural networks trained on FashionMNIST with MNIST samples yields influence functions that are roughly symmetrical --- the sample-sample influence matrix entry $M_{ij}$ is approximately equal to entry $M_{ji}$. This behavior is expected if the system is in good approximation in the kernel regime (when the neural tangent kernel prescribes symmetric generalization; see \S \ref{sec:ntk}).
This symmetry, however, is concerning because the influence of a bad, noisy sample on a good one is as big as the influence of a good sample on a bad one. As desired, the smaller convolutional models learn the most from similar stimuli and generalize less to further away stimuli.

Fine-tuning large language models, in contrast, yielded influence functions that are far more interesting, but disappointing. The pairwise influence matrix is now generally asymmetrical.
However, we find that this asymmetry does not reflect the types of inferential asymmetry we would like these models to possess~\cite{chen2024causal}: we found no signs of correct transitive behavior, respect for ontologies, causality, or compositionality. We found strong word dependence and weak meaning dependence.
In contrast, in-context processing (text production in the presence of relevant data in the same prompt), gets all the discussed desiderata of influence functions right.
Modern AI practitioners hope that their models can be fine-tuned, that the dynamics of learning from new data is such that it moves the resulting models into the right direction. If the resulting influence functions do reflect a good understanding of language then they should enable efficient model fine-tuning. Instead, our results suggest that the size of the fine-tuning datasets and context from the prompt somehow rescue performance in the face of influence functions that do not obey the basic desiderata. Our empirical influence function results suggest that popular models may be severely limited in their ability to learn the right things during fine-tuning.

\subsection{Influence functions in modern machine learning}

Influence functions have been used to study neural networks broadly, and LLMs in particular~\cite{zhao2024explainability,grosse_studying_2023}.
There are, however, known weaknesses of influence functions for evaluating neural networks.
In particular, influence functions that are computed from the gradients and Hessian of the model may grow fragile for sufficiently deep models~\cite{basu2020influence}.
Furthermore, these influence functions can have counterintuitive, even paradoxical interpretations:
For example, \citet{grosse_studying_2023} discover that the simple exposure of an autoregressive language model to a factual input (say, \textit{$A \in B$}) increases the likelihood of producing both the expected (say, \textit{$A \in B$}) as well as unexpected (say, \textit{$B \in A$}) outputs.

In order to better understand neural networks, we need to know how they reflect the actual way models use their training data to contextualize new knowledge domains, and if the unexpected behavior is a consequence of the influence function or of the learning paradigm itself.

\textbf{Key insight.} To see how elements of the fine-tuning set affect the resulting language model, we can directly read out the logits associated with a data point pre- vs post-finetuning. Some scientists choose a different approach that evaluates the internal states as well as the outputs \cite{grosse_studying_2023} --- usually driven by the desire to make statements about alignment or AI safety.
If we instead ignore internal state, this affords us a direct way of asking our questions: simply fine-tuning and observing the results.

\subsection{The neural tangent kernel}
\label{sec:ntk}

Under certain assumptions we can formulate the influences in a cleaner, mathematically far more appealing way. The neural tangent kernel (NTK) is one such formulation; it gives a good approximation to neural network learning when the model learning is locally well-approximated by a linear function. This realization shows how sometimes the notion of influence functions describing neural network behavior can be made very precise.

The NTK, \( \Theta(x, x') \), is defined as the inner product of the gradients with respect to the parameters, averaged over the random initialization of the parameters~\cite{domingos_every_2020}:

\begin{equation}
\Theta(x, x') = \mathbb{E}_{\theta} \left[ \nabla_\theta f(x; \theta)^\top \nabla_\theta f(x'; \theta) \right]
\end{equation}

The NTK describes the evolution of a neural network during learning. As the number of parameters approaches infinity, the NTK tends to stabilize, and recent work has indicated that LLMs (and indeed, even much smaller models) are sufficiently large that late-stage fine-tuning on many (but not all \cite{jacot2022freeze,hanin2019finite}) modern architectures live within the NTK regime~\cite{littwin2021random,malladi2023kernel,domingos_every_2020}.

Though the NTK cannot be experimentally materialized due to its (potentially infinite) size~\cite{jacot2018neural}, we can apply the computational limitations that are true of the NTK back to the original neural network.
The most relevant of these constraints is the fact that the NTK is a symmetric operator.  In other words, for training set $X_t$, $Y_t$ and test set $X_p$, $Y_p$, we compute \(\Delta W = -\eta \nabla_\theta \mathcal{L}(Y_t X_t)\).

We can approximate $\mathcal{L}_{W+\Delta W}(Y_p X_p)$ with the Taylor series expansion,

\begin{align}
\mathcal{L}_{W+\Delta W}(Y_p X_p) &= \mathcal{L}_W(Y_p X_p) + \Delta W^T \nabla_{\theta} (Y_p X_p) + \mathcal{O}(\Delta W^T \Delta W) \\
&\approx \mathcal{L}_W(Y_p X_p) - \eta \nabla_{\theta} \mathcal{L}(Y_t X_t)^T \nabla_{\theta} \mathcal{L}(Y_p X_p)
\end{align}

\textbf{Key insight.} The upshot of the NTK in this context is the intuition that the amount a sample $t$ informs the model about sample $p$ is the same magnitude by which sample $p$ informs the model about sample $t$.
In other words, $\mathcal{L}^t(p) = \mathcal{L}^p(t)$, where $\mathcal{L}^t$ represents the loss of the model fine-tuned on $t$.

The framework of empirical influence functions can be used more easily in the NTK regime of fine-tuning, where we can simply empirically train a model on a new domain (i.e., $\mathcal{P}$) and then demonstrate how a training sample in isolation affects the output.
Importantly, due to the speed of fine-tuning on small datasets, empirical influence functions can be efficiently calculated with the scale of compute available to most researchers.
This opens up the possibility of trying to understand the empirical influence functions instead of the overall input-output function, with the understanding that the influences may be simpler than the overall computation.
We may also hope that by analyzing influence functions we can obtain insights into the capabilities of modern deep learning systems, including questions about their ability to use basic logical and causal inference and how sparse the influence functions are.

\subsection{Empirical influence functions to understand fine-tuning}

Let us make our intuitions mathematically clear.
We consider the neural network model $f_\theta$ trained on loss $\mathcal{L}(Y X)$ through stochastic gradient descent (SGD) for training samples $X$ and labels $Y$.

We can begin with the universe of all knowledge, notated $\mathcal{T}$.
We then take a generalized model (say, a generalized, pretrained transformer, or GPT) that has been trained on all knowledge minus a previously-unseen data space, which we denote $\mathcal{T} \setminus \mathcal{P}$.
This model, notated $f_\theta^{\mathcal{T} \setminus \mathcal{P}}$, can be evaluated on datapoints $p \in \mathcal{P}$ to get the baseline sample loss, \(\mathcal{L}_{\theta}^{\mathcal{T} \setminus \mathcal{P}} (p)\).
We can then fine-tune on $\mathcal{P}$ to get a new ``complete'' model, and reevaluate those same points to get $\mathcal{L}_\theta^{\mathcal{T}}(p)$ for all $p \in \mathcal{P}$.

We can thus define our empirical influence function $\mathbf{EIF} (\mathcal{P})$ to be,

\begin{equation}
\mathbf{EIF} (\mathcal{P}) = \mathcal{L}_{\theta}^{\mathcal{T}} (p) - \mathcal{L}_{\theta}^{\mathcal{T} \setminus \mathcal{P}} (p) \text{~for~} p \in \mathcal{P}
\end{equation}

In order words, $\mathbf{EIF} (\mathcal{P})$ is the change in loss experienced by the model by learning the information in $\mathcal{P}$.
We highlight that the valence of this change can be positive \textit{or} negative (i.e., learning a few facts from $\mathcal{P}$ can enhance or detract from the model's ability to perform on $\mathcal{T} \setminus \mathcal{P}$).

Note that for this computation, we do not need to be aware of the specific weights involved in calculating the input-output function—only the inputs and logits; thus, our analyses are possible even on fine-tuneable black-box models.
We analyze both simple toy problems for which we have strong intuitions --- such as fine-tuning small CNNs --- and modern large language models, as we care about their interpretability and alignment.


\subsection{Contributions}

1. We focus on influence as opposed to the overall learned patterns, providing a conceptual understanding of fine-tuning with respect to model outputs.

2. We propose a set of desiderata for good neural network influence functions and find these desiderata to be violated.

3. We justify the notion that LLM prompt engineering can rescue behavior, even when learning violates the desiderata.

\section{Testing the approaches on FashionMNIST and MNIST}
\label{sec:mnist}

We first evaluate this technique on a simple domain for which our intuitions are straightforward.
First, let $\mathcal{T}$ be the domain of image space (specifically, the $28\times28$ grayscale pixel space).
Here, $\mathcal{T} \setminus \mathcal{P}$ is the domain of images comprising the FashionMNIST dataset~\cite{fashionmnist} and $\mathcal{P}$ is the set of digit images represented in the original MNIST dataset~\cite{deng2012mnist}.
Clearly, a model trained on $\mathcal{T} \setminus \mathcal{P}$ will have no prior knowledge whatsoever of $\mathcal{P}$, nor would we expect much knowledge to transfer between these domains.

We first train a simple PyTorch~\cite{torch,Falcon_PyTorch_Lightning_2019} convolutional model (CNN) $f_\theta^{\mathcal{T} \setminus \mathcal{P}}$ on the FashionMNIST dataset, and then fine-tune it on single examples $p \in \mathcal{P}$ from MNIST.
We also want to study the role of sample quality on this empirical influence function:
We will add progressive levels of pixelwise, iid, Gaussian noise $\mathcal{N}(0, 0.5)$ and $\mathcal{N}(0, 1)$ to simulate degraded samples, and compare the performance of the model on ``good'' as well as these ``noisy'' samples from $\mathcal{P}$.
A well-behaved model should learn less from low-quality samples than from high-quality samples,

\begin{equation}
\Delta\mathcal{L}_\theta^{p_i + \mathcal{N}(0, 1)}(p_j) \stackrel{?}{<} \Delta\mathcal{L}_\theta^{p_i}(p_j)
\end{equation}

This inequality has a corollary interpretation as an empirical influence function itself: In other words, we should like the influence of noise on the model to be null.

\begin{equation}
\mathbf{EIF}(\mathcal{N}(0, 1)) = \Delta\mathcal{L}_\theta^{p_i + \mathcal{N}(0, 1)}(p_j) - \Delta\mathcal{L}_\theta^{p_i}(p_j) \stackrel{?}{=} 0
\end{equation}

\subsection{Key results}

\begin{figure}
    \centering
    \includegraphics[width=0.75\linewidth]{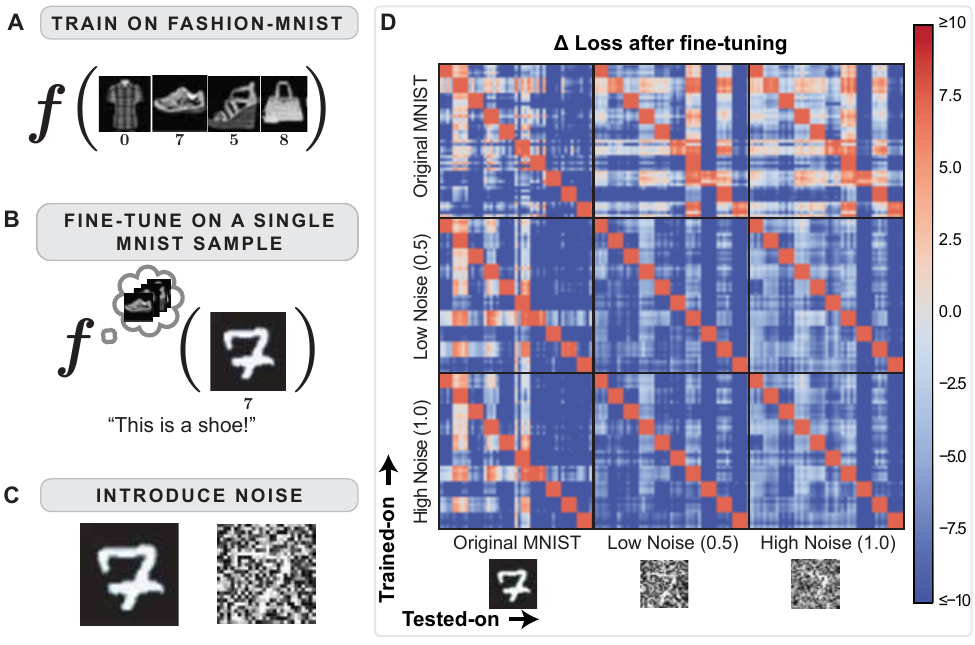}
    \caption{\small \textbf{Results of fine-tuning a CNN model on out-of-domain data.} \textbf{A. Train on Fashion-MNIST.} We first train a CNN model to correctly classify fashion items like shoes and purses. \textbf{B. Fine-tune on a single MNIST sample.} We then fine-tune that model on a \textit{single} example from the MNIST digit dataset. Here, we train the model to ``correctly'' classify the digit 7 as a shoe. \textbf{C. Introduce noise.} We add different levels of noise to simulate degraded data quality. \textbf{D. The pairwise EIF matrix.} We show a largely symmetric influence function. The blockwise pattern that emerges shows that digits are most informative about other digits of the same class.}
    \label{fig:mnist-results}
\end{figure}

We illustrate that the influence is symmetric --- the sample-sample influence matrix entry $\textbf{EIF}_{ij}$ is approximately equal to entry $\textbf{EIF}_{ji}$ (\textbf{Fig.~\ref{fig:mnist-results}}).
We also see a blockwise pattern emerge, because learning one digit dramatically increases the likelihood of correctly classifying another of the same digit. And model confusion becomes clear in this EIF as well (e.g., learning 2 influences 9 symmetrically as well, because of their visual similarity).

We find that noise has no influence on how much the model relies on a sample: we would like for the model to only learn from low-noise data but instead it does not distinguish (\(\textbf{EIF}(\mathcal{N}(0, 1)) \neq 0\)).

\section{Asymmetry in the Phi-3 LLM}

We next want to extend these findings to LLMs, a less-interpretable class of black box model, but with broad-reaching societal import.
Because LLMs are commonly trained on a broad universe of text corpora, organically finding a $\mathcal{P}$ --- a domain on which the LLM has not been trained --- is quite unlikely~\cite{Shi2023Detecting}.
Instead, we artificially construct a novel domain by producing structured facts from synthetically generated entity names (see \S\ref{sec:novelentity}).

Furthermore, we want to use models here that are accessible for general study.
Although parameter-efficient fine-tuning (PEFT) techniques like low-rank adapters (e.g., QLoRA~\cite{dettmers2023qlora}) are now widespread,~\cite{xiao2023smoothquant,wortsman2023stable} their quantitative differences to dense fine-tuning is still under study.
For this reason, we selected the Phi-3 model and here study the 3.82 billion-parameter model~\cite{abdin2024phi}.
Phi-3 is a relatively small LLM but with favorable reported performance characteristics on common benchmark logic evaluations~\cite{zhong2023agieval,hendrycks2020measuring,chollet2019measure,clark2019boolq}.

We begin with off-the-shelf pretrained weights which we denote $\mathcal{T}\setminus \mathcal{P}$.
We then construct $\mathcal{P}$ to contain both new domain knowledge as well as evaluation sequences that test the model against the desiderata we describe above.
We perform fine-tune training of a single epoch with a batch size of 1 on each $p \in \mathcal{P}$ to derive $n=|\mathcal{P}|$ fine-tuned models. We then evaluate each model on all other samples to compute the pairwise loss $\mathcal{L}_\theta^{p_i}(p_j)$ for $p_i, p_j \in \mathcal{P}$.
In comparison with the baseline model, the difference between the baseline loss $\mathcal{L}_\theta^{\mathcal{T} \setminus \mathcal{P}}(\cdot)$ and fine-tuned loss $\mathcal{L}_\theta^{\mathcal{T}}(\cdot)$ serves as our EIF.

Model loss for this LLM is the conditional log-probability of a token sequence.
To avoid influencing these probabilities with sequence length, we divide by the sequence length in token-count prior to computing the EIF:

\begin{equation}
\mathcal{L}_\theta = -\frac{1}{| N |} \sum_{i=1}^{N} \log P(t_i \mid t_1, t_2, \ldots, t_{i-1})
\end{equation}

\subsection{In-context operation rescues model failure}
\label{sec:prompting}

As LLM practitioners have now shown experimentally, LLMs perform much more effectively on logic and reasoning tasks when instructed to ``talk through'' the solution~\cite{kojima2022large,abdin2024phi}.
Furthermore, there is mounting evidence that in-prompt knowledge (often applied in retrieval-augmented generation, or RAG~\cite{lewis2020retrieval}) outperforms accessing the \textit{same information} from the training set, and that even minor changes to a prompt can have overwhelming influence on outputs~\cite{salinas2024butterfly}.
Using our empirical influence function approach, we can provide a mathematical basis for this phenomenon by comparing the EIF of fine-tuning on $\mathcal{P}$ with the EIF of providing samples from $\mathcal{P}$ in the prompt itself.

\subsection{Key results}

\begin{figure}
    \centering
    \includegraphics[width=0.9\linewidth]{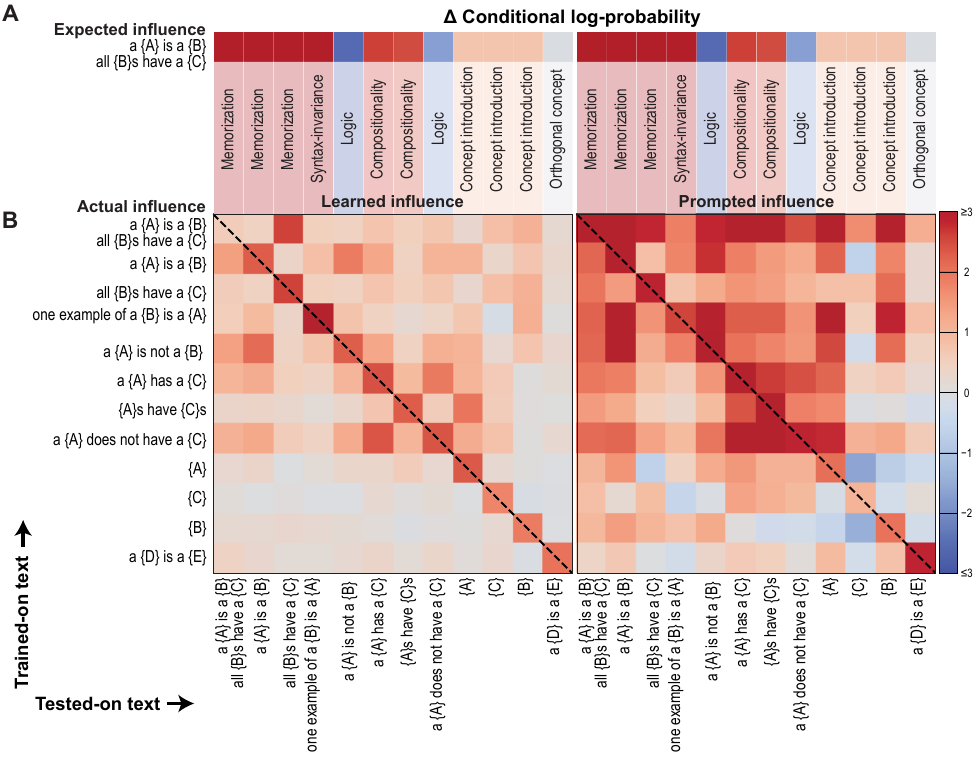}
    \caption{\small
    \textbf{The \texttt{transitivity} training example.}
    \textbf{A. Desired behavior for the first sample of training text in the matrix.}
    Here, we fine-tune a model on the two training samples, \textit{a \{A\} is a \{B\}} and \textit{all \{B\}s have a \{C\}}.
    We also show (vertical labels) the desiderata for which we are looking, per column.
    The obvious implication is that a \textit{\{A\}} should have a \textit{\{C\}} --- in other words, the $\Delta$conditional probability should increase.
    Paradoxically, the conditional probability of producing the tokens \textit{a \{A\} has a \{C\}} is barely affected, as is \textit{a \{A\} does not have \{C\}}.
    \textbf{B. The pairwise matrix of influence from training samples (vertical) to inference samples (horizontal) for learned EIF (left) and prompted EIF (right).}
    Horizontal ``banding'' patterns indicate that a training example uniformly influences other samples.
    Vertical banding indicates that an output tends to be made more likely when fine-tuning on any of the training samples.
    }
    \label{fig:phi3-ontology}
\end{figure}

\begin{figure}
    \centering
    \includegraphics[width=0.9\linewidth]{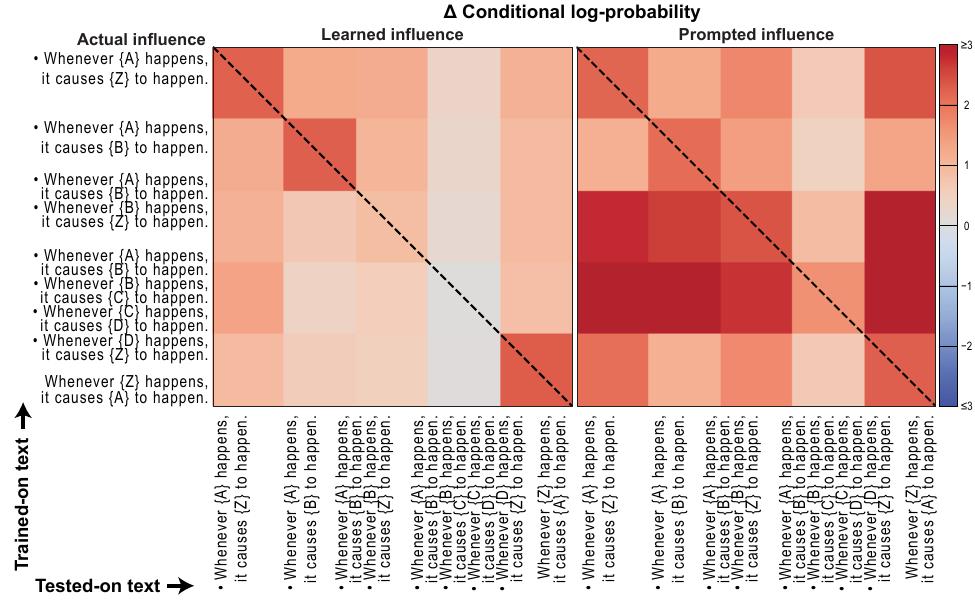}
    \caption{\small \textbf{Causal and logical induction in the \texttt{chain\_induces} training domain.} We would expect a learning machine to be able to describe the directionality of the causal arrow, but training on a chain of causes ($A\rightarrow B\rightarrow C\rightarrow D\rightarrow Z$, or even short subsections of this chain), makes it no more likely to produce the correct arrow from $A\rightarrow Z$, versus the incorrect arrow, $Z\rightarrow A$. Providing this information in the prompt rather than in training samples (right pane) rescues the model's performance.}
    \label{fig:phi3-induction}
\end{figure}

\begin{figure}
    \centering
    \includegraphics[width=0.9\linewidth]{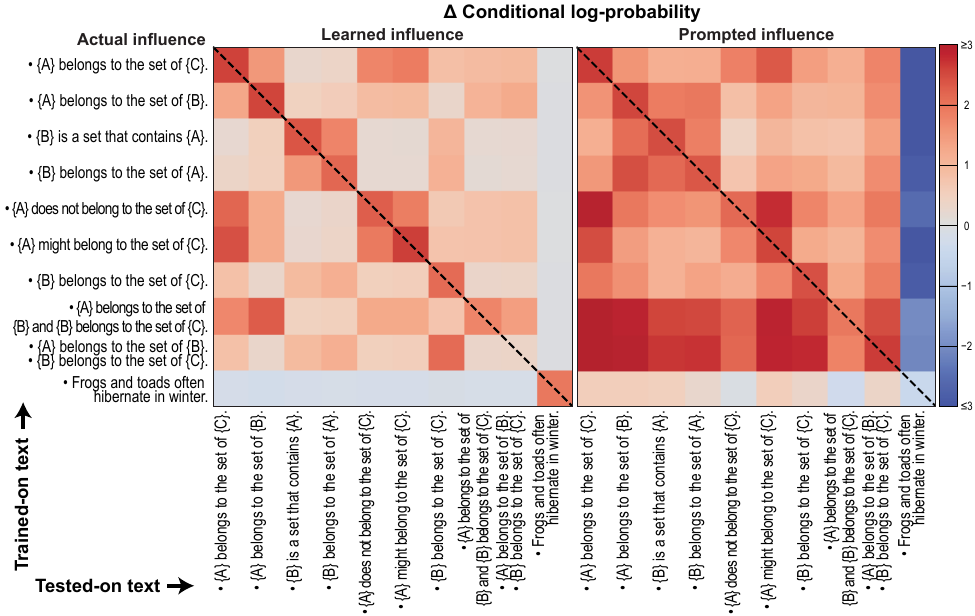}
    \caption{\small
    \textbf{Ontological reasoning in the \texttt{belongs\_to} training domain.}
    Set containment is an asymmetric operator: Not all rectangles are squares. But in the absence of prior knowledge of a domain, learning $A\in B$ does not endow an LLM with the ability to distinguish between $B \ni A$ and $B \in A$.
    An out-of-domain sample (``\textit{Frogs and toads often hibernate in winter.''}) is provided to illustrate that orthogonal examples are affected (the EIF shown in the last column and row are slightly negative).
    On the right, we illustrate that including information in the prompt rather than in training samples partially rescues the model's performance on ontological reasoning, but token order is unreasonably important.
    }
    \label{fig:phi3-onto}
\end{figure}

LLM fine-tuning revealed influence functions that are generally asymmetrical.
This is compatible with what we would normatively hope, as asymmetry in this EIF is required to encode things like causal or logical implication.
But instead, our analysis of pairwise influences revealed that this asymmetry does not reflect the types of inferential asymmetry we would like these models to possess:
There is, for example, no stronger influence of the notion ``dogs $\in$ mammals'' on the output ``my dog $\in$ mammals'' than ``mammals $\in$ my dog.'' The analyzed model thus does not carry the kind of asymmetry we would ideally want to have in such cases.

As shown in \textbf{figures~\ref{fig:phi3-ontology}, \ref{fig:phi3-induction}, and \ref{fig:phi3-onto}}, the tendency to a symmetric EIF --- as induced by the symmetric NTK operator --- can be overcome by introducing information in the prompts.
In other words, \textit{training} on a concept that requires inferential or implied asymmetry results in a symmetric EIF, whereas \textit{prompting} on an asymmetric concept results in a likewise \textit{asymmetric} EIF.

\textbf{Transitivity.} We found that the model was unable to perform transitive logic when provided with chained training samples:
Training on the separate training samples $\{$\textit{``A is bigger than B''}; \textit{``B is bigger than C''}$\}$ led to equal likelihood of producing \textit{``A is bigger than C''} and \textit{``C is bigger than A''}.
This failure could be partially rescued by training on the combined (single) training sample, $\{$\textit{``A is bigger than B,~B is bigger than C''}$\}$, though this remedy depends on modifications to the training set.

\textbf{Ontology \& Compositionality.} When presented with the separate training samples $\{$\textit{``A $\in$ B''}; \textit{``B $\in$ C''}$\}$, the model was not more likely to produce \textit{``A $\in$ C''} than \textit{``A $\notin$ C''} nor \textit{``C $\in$ A''}, compatible with the findings from \citet{grosse_studying_2023}. (Our natural-language equivalents of these training samples are described in \textit{Supplemental Material}.)

\textbf{Causality.} We found no evidence that these models can carry logic or causality through chained statements in the fine tuning set (i.e., training on $\{$``\textit{A implies B}''; ``\textit{B implies C}''$\}$ has no meaningful influence on the output ``\textit{A implies C}''), compatible with the findings from \citet{chen2024causal}.

\begin{figure}
    \centering
    \includegraphics[width=0.9\linewidth]{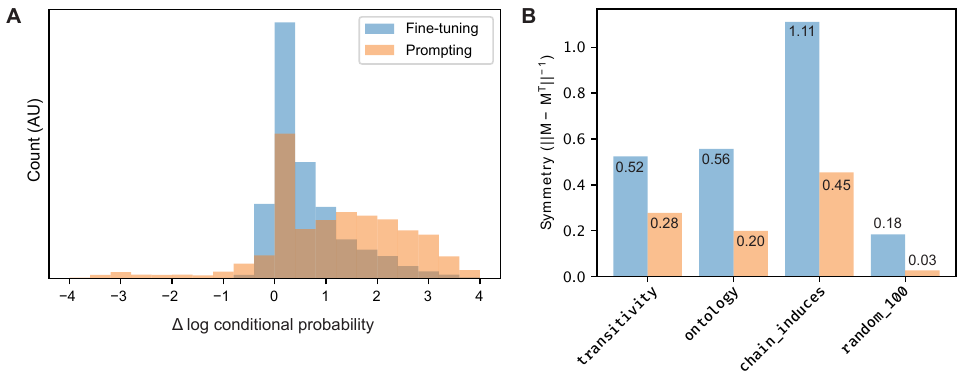}
    \caption{\small
    \textbf{A. Histogram of empirical influence function values, showing diffusivity of influence.}
    The distribution of EIF values for learned knowledge is much more diffuse and positive-valued than that of prompted, in-context processing. We interpret this to suggest that models pull both relevant and irrelevant information when formulating a response, rather than retrieving specific, relevant information, as we would hope. Note the negative tail of the distribution, unique to the prompting condition. The histogram is shown for all synthetic knowledge domains. Individual histograms per domain are reported in \textit{Supplemental Material}. (AU = arbitrary units.)
    \textbf{B. Symmetry measure breakdown per domain.} For \textit{all} knowledge domains, prompting leads to a more asymmetric EIF than fine-tuning.
    }
    \label{fig:diffusivity}
\end{figure}

\textbf{Sparsity.} The empirical influence functions of all $\mathcal{P}$ were densely populated (\textbf{Fig.~\ref{fig:diffusivity}}).
In other words, the model did not use relevant knowledge more than it used irrelevant or even contrary data.
This is in stark contrast to the logical relations between sentences in human-produced text.

In a way, we can think of LLMs as setting up influence functions which then define how fine-tuning data affects model performance. In contrast, in-context processing (text production in the presence of relevant data in the same prompt) gets more of the discussed desiderata of influence functions right (see \S\ref{sec:prompting}).
Modern AI practitioners hope that influence functions reflect a good understanding of language, but we instead propose that the size of the dataset and prompt context rescues performance despite bad model generalization.
Instead, our empirical influence function results suggest popular models cannot generalize or perform logic in the way they appear to.

\subsection{Novel knowledge domain generation}
\label{sec:novelentity}

In order to produce a novel knowledge domain, we generated fact statements with new words unlikely to be in the training set.
We formed realistic words by randomly sampling from a set of compatible syllables to form phonetically reasonable entity names (i.e., ``\textit{klarbu}'' but not ``\textit{klrb}'').
Note that this name generation has \textit{no} correlation with token prevalence or $n$-grams in the language model or tokenizer in use, as this would run the risk of preferring entity names that exist in the training set.
Our word generation code is included in the \textit{Supplemental Materials}.

We then constructed small knowledge domains to interrogate the model for our desiderata described above (see \textbf{Figs.~\ref{fig:phi3-ontology}, \ref{fig:phi3-induction}, and \ref{fig:phi3-onto}}).
For example, to study the directivity of ontology, we can construct a simple template, \textit{a \{A\} is a \{B\}}. We can then randomly sample from the universe of synthetic words to populate the \textit{A} and \textit{B} variables and generate the training sample, e.g., \textit{``a \textbf{klarbu} is a \textbf{pimjon}''}.
We randomly generate a large number of complete knowledge domains to avoid the coincidental generation of a word that evokes a real-world meaning that might occur in the training set.
Results in this section are reported for $n=10$ repeat trials, averaged.

\subsection{Discovering a learning rate for a training set with EIF}

Correctly calibrating a learning rate $\eta$ for fine-tuning requires balancing learning efficiency with instability.
This challenge can be easily addressed by a simple parameter sweep of $\eta$ during fine-tuning.
Values that are too low will result in no change to $\mathcal{L}$ at all ($\textbf{EIF}=0$).
Values that are too high will result in an EIF where noisy variance dominates signal.
Between these, there is a stable plateau from which $\eta$ can be safely selected.
This research uses $\eta=1\times 10^{-6}$ for all LLM fine-tuning.

\section{Discussion}
Here, we measured empirical influence functions of several neural networks under different fine-tuning conditions, and compared these results to a list of desiderata for these models based upon popular assumptions about LLM behavior. We found no evidence of asymmetry with uncertainty: systems learn as much from bad data and generalize it to good data as they learn from good data and generalize to bad data. There is no sign of in-memory chaining, nor of implication, set membership, causality, or compositionality. There is also massive variance in the influence functions driven by irrelevant features, e.g. the specific (new) words used. Even the simple concept of learning more strongly from more similar stimuli (e.g. the same stimulus) is often violated. In summary, influence functions of fine-tuning (but not prompting) appear remarkably bad even though overall obtained trained models after fine-tuning are known to appear quite good.

Our approach to understanding fine-tuning in deep learning is based on linearization. The local linearity assumption is going to be good if weight changes are small relative to the involved nonlinearities. It is particularly meaningful for wide neural networks (often given, \cite{jacot2018neural}), that are not too deep (questionable, although skip-connections often render them less deep in practice \cite{jacot2022freeze}), and weights that are relatively large~\cite{jacot2018neural}. For fine-tuning, the relevance of the linear regime has been verified for a range of LLMs \cite{hanin2019finite,jacot2022freeze}. However, our empirical influence function approach is relevant beyond that. Regardless of the linearity of the system, for fine-tuning to be useful, it should follow the desiderata we have for it. Future research is needed to disentangle the relationship of the linear regime relative to the empirical influence functions we studied here.

Our approach is relatively slow: for every training stimulus, it requires one or a handful of gradient descent steps, and then for every test stimulus it requires an inference calculation. Our method thus has wallclock-time characteristics of fine-tuning a new model (on approximately one new sample) in under one second, and computing our pairwise EIF matrix is trivially $\mathcal{O}(|\mathcal{P}|^2)$. This is fast enough to perform many fine-tuning experiments, but we clearly could not compute this EIF for millions of samples, or anything close to the scale of $|\mathcal{T}|$. At least in the linear regime, work along the lines of \cite{grosse_studying_2023} would allow us to scale up these experiments.

Our desiderata were also somewhat superficial, reflecting only high-level normative expectations for good learning system. Linguists have worked extensively on the rich ways how human language is structured and the kind of logical relations that it often exhibits. It would thus make a lot of sense to ground work like ours more deeply in an understanding of human communication.

This work gives us great insight into learning. In this view, fine-tuned ANNs are basically lookup machines~\cite{domingos_every_2020} that do not think or reason, but instead utilize influence functions ~\cite{zheng2023intriguing,feldman2020neural} that do not obey practitioners' desiderata. If we are dealing with lookup machines, we should at least demand the influences used for this lookup obey the general principles we have for knowledge ontologies. This is particularly problematic because we want to fine-tune for use cases that are meaningfully outside of our data distribution. Focusing on influence opens a new way of understanding fine-tuning and its failure modes.

Logical chaining works within the prompt but not between elements learned in the past. Sometimes this is no problem because multiple pieces of information appear in the same context. However, this failure mode may partially explain the usefulness of retrieval augmented generation, which moves relevant information into the context. A stronger, maybe more brain-like, way of pulling relevant information from earlier in the training set (episodic memory) into the context may thus be seen as a potential strategy to ameliorate the violations of the desiderata we discussed.

The failure to fine-tune well, as we have characterized here, may be relevant beyond the domain of fine-tuning. Fine-tuning is, after all, just the very last moments of learning. This work should be extended to determine if the process of LLM training broadly requires significant revision.

Our focus on influence here promises a straightforward way of analysing and improving deep learning models. We want their influences to exhibit the desiderata above. This could directly be built into meta-learning systems that optimize the influences instead of optimizing the input-output behavior. This should be doable directly --- we remind the reader that our influence calculations are differentiable. Getting closer to the desiderata may be possible with simple strategies, e.g., a low-dimensional way of regulating learning speed and location. Efficient ways to implement such meta-learning would be a useful endeavor for the future of deep learning.

\begin{ack}
Research in this publication was supported by the National Institutes of Health under award number UC2-NS128361. 
The content is solely the responsibility of the authors and does not necessarily represent the official views of the National Institutes of Health.
\end{ack}


\medskip

{
\bibliographystyle{plainnat}
\bibliography{references}

}

\clearpage
\appendix

\section{Appendix / supplemental material}



\subsection{Histograms of influence per-domain}

\begin{figure}
        \centering
        \includegraphics[width=1\linewidth]{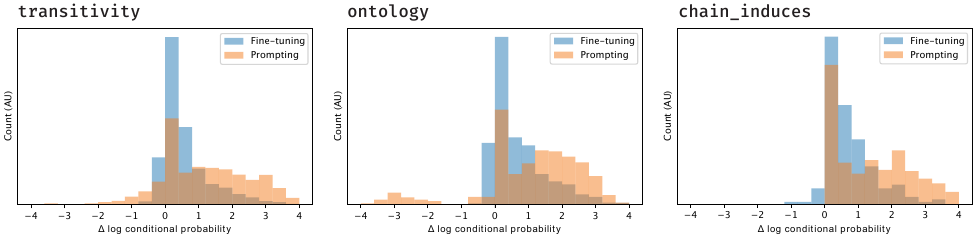}
        \caption{Histograms of pairwise influence function matrices for each synthetic knowledge domain constructed in this study.}
        \label{fig:histos}
\end{figure}

Here, we report the histograms of EIF values for each synthetic knowledge domain reported (\textbf{Fig.~\ref{fig:histos}}).
These are averaged across 10 random selections of entity name generation.

\subsection{Synthetic knowledge domains}

\subsubsection{belongs\_to}
\begin{itemize}
\item
\begin{itemize}
\item \texttt{\{A\}} belongs to the set of \texttt{\{C\}}.
\end{itemize}
\item
\begin{itemize}
\item \texttt{\{A\}} belongs to the set of \texttt{\{B\}}.
\end{itemize}
\item
\begin{itemize}
\item \texttt{\{B\}} is a set that contains \texttt{\{A\}}.
\end{itemize}
\item
\begin{itemize}
\item \texttt{\{B\}} belongs to the set of \texttt{\{A\}}.
\end{itemize}
\item
\begin{itemize}
\item \texttt{\{A\}} does not belong to the set of \texttt{\{C\}}.
\end{itemize}
\item
\begin{itemize}
\item \texttt{\{A\}} might belong to the set of \texttt{\{C\}}.
\end{itemize}
\item
\begin{itemize}
\item \texttt{\{B\}} belongs to the set of \texttt{\{C\}}.
\end{itemize}
\item
\begin{itemize}
\item \texttt{\{A\}} belongs to the set of \texttt{\{B\}} and \texttt{\{B\}} belongs to the set of \texttt{\{C\}}.
\end{itemize}
\item
\begin{itemize}
\item \texttt{\{A\}} belongs to the set of \texttt{\{B\}}.
\item \texttt{\{B\}} belongs to the set of \texttt{\{C\}}.
\end{itemize}
\item
\begin{itemize}
\item Frogs and toads often hibernate in winter.
\end{itemize}
\end{itemize}
\subsubsection{chain\_induces}
\begin{itemize}
\item
\begin{itemize}
\item Whenever \texttt{\{A\}} happens, it causes \texttt{\{Z\}} to happen.
\end{itemize}
\item
\begin{itemize}
\item Whenever \texttt{\{A\}} happens, it causes \texttt{\{B\}} to happen.
\end{itemize}
\item
\begin{itemize}
\item Whenever \texttt{\{A\}} happens, it causes \texttt{\{B\}} to happen.
\item Whenever \texttt{\{B\}} happens, it causes \texttt{\{Z\}} to happen.
\end{itemize}
\item
\begin{itemize}
\item Whenever \texttt{\{A\}} happens, it causes \texttt{\{B\}} to happen.
\item Whenever \texttt{\{B\}} happens, it causes \texttt{\{C\}} to happen.
\item Whenever \texttt{\{C\}} happens, it causes \texttt{\{D\}} to happen.
\item Whenever \texttt{\{D\}} happens, it causes \texttt{\{Z\}} to happen.
\end{itemize}
\item
\begin{itemize}
\item Whenever \texttt{\{Z\}} happens, it causes \texttt{\{A\}} to happen.
\end{itemize}
\end{itemize}

\subsubsection{transitivity}
\begin{itemize}
\item
\begin{itemize}
\item a \texttt{\{klarbu\}} is a \texttt{\{pimjon\}}
\item all \texttt{\{pimjon\}}s have a \texttt{\{greev\}}
\end{itemize}
\item
\begin{itemize}
\item a \texttt{\{klarbu\}} is a \texttt{\{pimjon\}}
\end{itemize}
\item
\begin{itemize}
\item all \texttt{\{pimjon\}}s have a \texttt{\{greev\}}
\end{itemize}
\item
\begin{itemize}
\item one example of a \texttt{\{pimjon\}} is a \texttt{\{klarbu\}}
\end{itemize}
\item
\begin{itemize}
\item a \texttt{\{klarbu\}} is not a \texttt{\{pimjon\}}
\end{itemize}
\item
\begin{itemize}
\item a \texttt{\{klarbu\}} has a \texttt{\{greev\}}
\end{itemize}
\item
\begin{itemize}
\item \texttt{\{klarbu\}}s have \texttt{\{greev\}}s
\end{itemize}
\item
\begin{itemize}
\item a \texttt{\{klarbu\}} does not have a \texttt{\{greev\}}
\end{itemize}
\item
\begin{itemize}
\item \texttt{\{klarbu\}}
\end{itemize}
\item
\begin{itemize}
\item \texttt{\{greev\}}
\end{itemize}
\item
\begin{itemize}
\item \texttt{\{pimjon\}}
\end{itemize}
\item
\begin{itemize}
\item a \texttt{\{skimoy\}} is a \texttt{\{krombom\}}
\end{itemize}
\end{itemize}

\subsubsection{squares}
\begin{itemize}
\item
\begin{itemize}
\item SQUARE belongs to the set of \texttt{\{B\}}.
\end{itemize}
\item
\begin{itemize}
\item \texttt{\{B\}} belongs to the set of RECTANGLE.
\end{itemize}
\item
\begin{itemize}
\item SQUARE belongs to the set of \texttt{\{B\}} and \texttt{\{B\}} belongs to the set of RECTANGLE.
\end{itemize}
\item
\begin{itemize}
\item SQUARE belongs to the set of \texttt{\{B\}}.
\item \texttt{\{B\}} belongs to the set of RECTANGLE.
\end{itemize}
\item
\begin{itemize}
\item \texttt{\{B\}} belongs to the set of SQUARE.
\item \texttt{\{B\}} belongs to the set of RECTANGLE.
\end{itemize}
\item
\begin{itemize}
\item RECTANGLE belongs to the set of SQUARE.
\end{itemize}
\end{itemize}

\subsubsection{Random sentence domains}

The \texttt{random\_claim} and \texttt{random\_claim\_100} domains, which are constructed from ArXiv preprint abstracts, are described further (with provenance) in the codebase.

\subsection{Synthetic word design}

Words for our synthetic knowledge domains were selected by randomly shuffling and selecting between two and five items from our \textit{syllables} list (shown in the attached codebase under \textit{sample\_generator.py $\rightarrow$ create\_variable}).
Spot-checks confirmed that these words tended not to exist in the training set.
We also took care to rerun each experiment multiple times with different randomization seeds (reported in the attached codebase) in order to reduce the possibility of a particular synthetic word influencing our results.

We deliberately avoided the use of tokens or other statistical information from the LLM preprocessors and tokenizers, as this would inadvertently increase the likelihood of producing words that the model had seen in its training set.

\end{document}